\documentclass[runningheads]{llncs}

\usepackage[T1]{fontenc}
\usepackage{graphicx}
\usepackage{hyperref}
\usepackage{color}

\usepackage{amsmath}
\usepackage{amssymb}
\usepackage{bm}
\usepackage{xurl}

\usepackage[utf8]{inputenc}
\usepackage{multirow}
\usepackage{booktabs}
\usepackage{graphicx}
\usepackage{array}
\usepackage{subfig}
\usepackage{fancyhdr}
\begin{document}

\title{Impact of Data Augmentation on Confidence Calibration in Melanoma Classification}
\titlerunning{Impact of Augmentation on Calibration in Melanoma Classification}
% If the paper title is too long for the running head, you can set an abbreviated paper title here
\author{Morgan May\inst{1,4}\orcidID{0009-0006-1953-7600} \and
Simon Caton\inst{2}\orcidID{0000-0001-9379-3879} \and
Pierpaolo Dondio\inst{3}\orcidID{0000-0001-7874-8762}}
\authorrunning{Morgan May et al.}
\institute{Technological University Dublin, Dublin, Ireland \and
University College Dublin, Dublin, Ireland \and
Dublin City University, Dublin, Ireland \and
Research Ireland Centre for Research Training in Machine Learning, ML-Labs, Dublin, Ireland\\
\email{simon.caton@ucd.ie}}

\maketitle
\thispagestyle{fancy} % Forces the right-anchored header onto the first page
\begin{abstract}
Accurately quantifying the predictive uncertainty or improving model calibration plays an important role in medical image classification, 
in particular in melanoma diagnosis, where accurate uncertainty quantification can have significant implications for patient care.
One of the methods for calibration improvement is data augmentation. 
In addition, data augmentation as a method for synthetically increasing the size of the dataset has been proven to improve the performance of models trained on imbalanced datasets.
However, the impact of data augmentation, as a transformation of a part of the original data, on calibration of models trained on imbalanced datasets, in particular in melanoma classification is under-explored.
We train neural networks on SIIM-ISIC 2020 melanoma classification dataset under two conditions: with and without data augmentation, and compare the differences in AUC and expected calibration error (ECE) in both scenarios. Our results shows improvements in uncertainty calibration using different augmentation methods.
\end{abstract}

\section{Introduction}
Skin cancer is reported to be the most common form of cancer in Ireland, with approximately 7,000 annual new cases \cite{stjames_melanoma}. Even the most dangerous form of skin cancer, malignant melanoma can be cured, if diagnosed in early stages.

Deep learning has transformed medical image classification, for example, in melanoma diagnosis \cite{esteva2017dermatologist,meedeniya2024skin,naseri2025diagnosis}.
Recently, the reliability of the confidence of model predictions, referred to as calibration, has attracted attention \cite{guo2017calibration,minderer2021revisiting}. 
This is essential especially for safety sensitive applications such as medical diagnosis \cite{lambert2024trustworthy,lopez2025uncertainty,wahid2024artificial}, and specifically in melanoma classification \cite{asgharnezhad2026quad,kim2025melanoma,maron2022model}. Calibration refers to the alignment between predicted probabilities and observed event frequencies.
Efforts to improve the calibration of neural networks have led to the development of various methods \cite{fakour2024structured,wang2025aleatoric}. 

Data augmentation has shown benefits in calibration improvement \cite{cao2024survey,jin2024survey,maronas2021calibration,zhang2022and}.
It increases data diversity by applying different transformations to the original data samples. It is
also used to increase the training samples \cite{shorten2019survey}, in particular for skin lesion diagnosis \cite{patil2025enhancing,perez2018data,wu2022skin}. Test-time augmentation is shown to improve the calibration of different neural networks trained on melanoma dataset under distribution shift \cite{hekler2023test}.
Authors in \cite{rao2023studying} evaluated the effect of mixup \cite{zhang2018mixup}, CutMix \cite{yun2019cutmix}, and cutout \cite{devries2017improved} on the calibration and performance of ResNet-50 and ResNet-101 for medical tasks. They report that in terms of AUC and accuracy, ResNet-50 and ResNet-101 achieved no significant performance improvement or even worse performance by using these augmentations, while they all improved the calibration. This work evaluates the impact of a wider range of augmentation methods including mixup and CutMix from mixing augmentation family, and trivialaugment and RandAugment from automated augmentation family, on EfficientNet-B0 and EfficientNet-B2 \cite{tan2019efficientnet} that have shown state of the art performance on ISIC melanoma data \cite{ha2020identifying}. We examine the performance of the models on melanoma classification by using data augmentation as a partial transformation to a randomly selected part of the training data.
Our experiments show that partial data augmentation can improve expected calibration error (ECE), while maintaining a competitive AUC, or even improving AUC slightly.

\section{Methodology} \label{sec:methodology}
\vspace{-0.15in}
This work explores the effects of augmentation on the calibration of the models on melanoma classification. As such, we do not intend to compare the models against each other or optimise the performance of the models. Therefore, we did not include any additional steps such as ensembling.

\noindent \textbf{Dataset.}
We use the SIIM-ISIC 2020 Melanoma Classification dataset \cite{isic2020dataset}, which is imbalanced with an imbalance ratio of about 55. We used the 33126 labeled images in the original trainset for a train/validation/test split of 80/10/10.

\noindent \textbf{Models.}
We train two EfficientNets \cite{tan2019efficientnet},
EfficientNet-B0 and EfficientNet-B2. The models are pretrained on ImageNet.

\noindent \textbf{Data Augmentation.} We use two mixing augmentation methods, mixup \cite{zhang2018mixup} and CutMix \cite{yun2019cutmix}, and two automated augmentation methods, trivialaugment \cite{muller2021trivialaugment} and RandAugment \cite{cubuk2020randaugment}, for partial transformation of a random selection of training examples.

\noindent \textbf{Loss Function.}
For the training on the %highly 
imbalanced dataset, we use the \textit{logit adjusted softmax cross-entropy loss} \cite{menon2021longtail}. 
To compute the logit adjusted softmax cross-entropy loss, we shift the logits by $\tau \log(\pi_y)$, where $\pi_y$ is the class prior probability and $\tau$ is a hyperparameter known as temperature.

\noindent \textbf{Performance metrics.}
We use AUC for the performance evaluation, and expected calibration error (ECE) \cite{guo2017calibration,naeini2015obtaining} for calibration evaluation. 

\section{Results and Discussion} \label{sec:eval}
The evaluation metrics AUC and ECE for EfficientNet-B0 and EfficientNet-B2 using different augmentation methods are shown in \tablename~\ref{tab:summary}. 

\noindent \textbf{Calibration.}
The baseline ECE values for EfficientNet-B0 and EfficientNet-B2 are 0.0500 and 0.0207, respectively. This means that EfficientNet-B2 has a better uncertainty quantification than EfficientNet-B0. All augmentations show improvements in ECE for both models, with the highest improvement of 0.0321 in ECE from the baseline by using trivialaugment for EfficientNet-B0, and with the highest improvement of 0.0039 in ECE from the baseline by using trivialaugment for EfficientNet-B2.

\noindent \textbf{Performance.}
For EfficientNet-B0, mixup, RandAugment and trivialaugment performed slightly worse in terms of AUC, while CutMix improved AUC slighly. For EfficientNet-B2, CutMix, mixup and RandAugment performed slightly worse in terms of AUC, while trivialaugment improved AUC slighly.

\vspace{-0.2in}

\begin{table}[]
\caption{Impact of Data Augmentation on Melanoma Classification}
\centering
\begin{tabular}{|c|cc|cc|}
\hline
\multirow{2}{*}{\textbf{Augmentation}} & \multicolumn{2}{c|}{\textbf{EfficientNet-B0}}    & \multicolumn{2}{c|}{\textbf{EfficientNet-B2}}    \\ \cline{2-5} 
                                       & \multicolumn{1}{c|}{\textbf{AUC}} & \textbf{ECE} & \multicolumn{1}{c|}{\textbf{AUC}} & \textbf{ECE} \\ \hline
None                                      & \multicolumn{1}{c|}{0.8808}       & 0.0500       & \multicolumn{1}{c|}{0.8836}       & 0.0207       \\ \hline
CutMix                                 & \multicolumn{1}{c|}{\textbf{0.8937}}       & 0.0263       & \multicolumn{1}{c|}{0.8823}       & 0.0179       \\ \hline
mixup                                  & \multicolumn{1}{c|}{0.8787}       & 0.0254       & \multicolumn{1}{c|}{0.8695}       & 0.0198       \\ \hline
RandAugment                            & \multicolumn{1}{c|}{0.8655}       & 0.0205       & \multicolumn{1}{c|}{0.8762}       & 0.0179       \\ \hline
trivialaugment                         & \multicolumn{1}{c|}{0.8802}       & \textbf{0.0179}       & \multicolumn{1}{c|}{\textbf{0.8843}}       & \textbf{0.0168}       \\ \hline
\end{tabular}
\label{tab:summary}
\end{table}

\vspace{-0.3in}
\section{Conclusion} \label{sec:conclusion}
We evaluated the effect of the partial application of data augmentation on the uncertainty calibration of deep learning models in melanoma classification.
Our aim is to compare the calibration of models in two scenarios: with- and without-augmentation. 
Partial application of different data augmentations to the training data has been shown to improve the calibration of models trained on SIIM-ISIC melanoma classification dataset. 
This means that if data augmentation is applied to a part of the melanoma dataset, the model will produce prediction probabilities that are more similar to the actual outcomes of an image representing either a malignant case or a benign case, which leads to safer melanoma diagnosis.
It will be interesting to see the effect of a greater variety of augmentation methods on a wider range of models in melanoma classification. 

\section{Acknowledgment}
This publication has emanated from research conducted with the financial support of Taighde Éireann – Research Ireland under Grant number 18/CRT/6183. 
For the purpose of Open Access, the author has applied a CC BY public copyright licence to any Author Accepted Manuscript version arising from this submission. 

\bibliographystyle{splncs04}
\bibliography{Bibliography}
\end{document}